\documentclass[lettersize,journal]{IEEEtran}
\usepackage{amsmath,amsfonts}
\usepackage{algorithmic}
\usepackage{algorithm}
\usepackage{array}
\usepackage[caption=false,font=normalsize,labelfont=sf,textfont=sf]{subfig}
\usepackage{textcomp}
\usepackage{stfloats}
\usepackage{url}
\usepackage{verbatim}
\usepackage{graphicx}
\usepackage{cite}
\usepackage{booktabs}
\usepackage{multirow}
\usepackage{adjustbox}

\usepackage[table]{xcolor}
\definecolor{bestred}{RGB}{248,224,224}
\definecolor{deepblue}{RGB}{0, 0, 180}
\definecolor{deepgreen}{RGB}{0, 110, 70}
\definecolor{urlpurple}{RGB}{140, 40, 130}

\usepackage[colorlinks=true, linkcolor=deepblue, citecolor=deepgreen, urlcolor=urlpurple]{hyperref}

\usepackage{pifont}

\begin{document}

\title{DeblurNVS: Geometric Latent Diffusion for Novel View Synthesis from Sparse Motion-Blurred Images}

\author{Changyue Shi, Wangbo Yu, Chaoran Feng, Li Yuan
\thanks{Changyue Shi is with the School of AI for Science, Peking University Shenzhen Graduate School, Shenzhen, 518055 China (Email: changyue\_shi@163.com)}
\thanks{Wangbo Yu is with the School of Electronic and Computer Engineering, Peking University Shenzhen Graduate School, Shenzhen, 518055 China (Email: yuwangbo98@gmail.com)}
\thanks{Chaoran Feng is with the School of Electronic and Computer Engineering, Peking University Shenzhen Graduate School, Shenzhen, 518055 China (Email: chaoran.feng@stu.pku.edu.cn)}
\thanks{Li Yuan is with the School of AI for Science and the School of Electronic and Computer Engineering, Peking University Shenzhen Graduate School, Shenzhen, 518055 China (Email: yuanli-ece@pku.edu.cn)}

}



\maketitle

\begin{abstract}
{Novel view synthesis (NVS) is a fundamental problem in computer vision and graphics. Recent advances in neural radiance fields (NeRF), 3D Gaussian Splatting (3DGS), and generative view synthesis have substantially improved its quality.} {Yet most methods still rely on clean observations, where image structures and cross-view geometric cues are well preserved.} {Motion blur breaks this assumption by corrupting local details and weakening multi-view correspondences. Such blur commonly arises from camera shake, scene motion, or finite exposure in practical capture.} {Blur-aware NVS methods address this degradation by modeling image formation, but their reliance on costly per-scene optimization limits efficient and generalizable sparse-view synthesis.} To address this, we propose DeblurNVS, a novel framework for synthesizing high-fidelity novel views directly from sparse motion-blurred images, without requiring {per-scene optimization}. {DeblurNVS restores the intermediate geometric representations needed for multi-view reasoning, enabling blurred inputs to recover reliable structure and correspondence cues.} {The restored representations are then combined with target camera information to synthesize the target-view representation and reconstruct a sharp RGB novel view.} To enable the large-scale training, we construct a motion-blurred NVS dataset from DL3DV-10K using interpolation-based finite-exposure blur synthesis.
Extensive experiments demonstrate that DeblurNVS {outperforms existing baselines on synthetic motion-blur benchmarks and generalizes to real motion-blurred scenes, producing perceptually sharper and structurally more stable novel views while avoiding costly per-scene optimization}. Project page: \url{https://github.com/PKU-YuanGroup/DeblurNVS}.

\end{abstract}


\section{Introduction}

\begin{figure}
\centering
\includegraphics[width=0.5\textwidth]{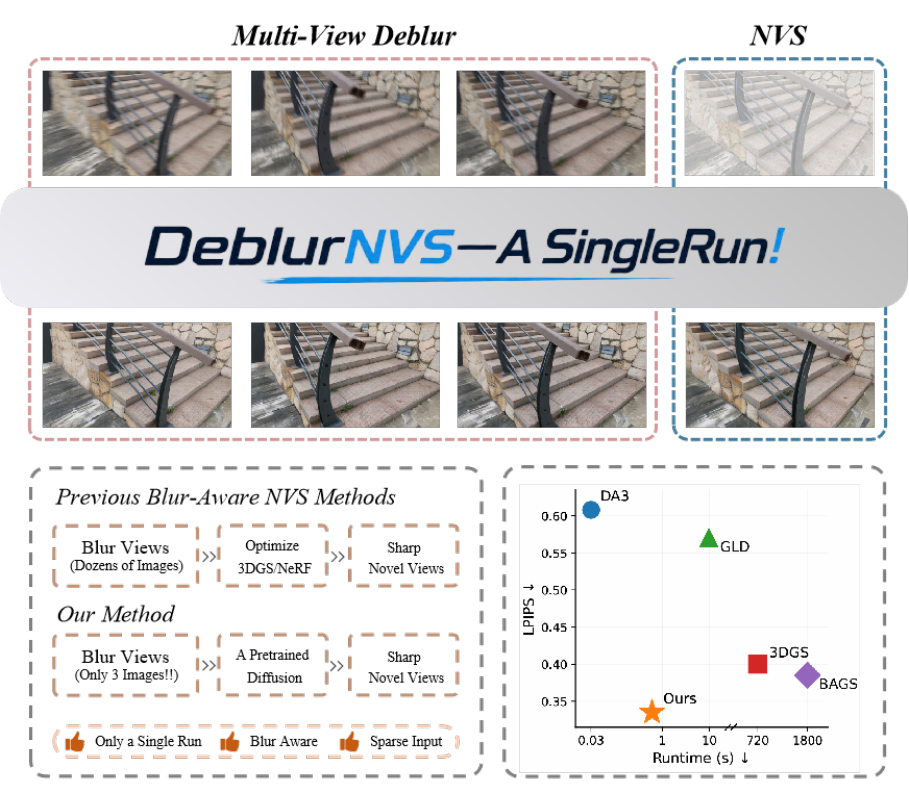}
\caption{We propose DeblurNVS, a novel framework that synthesizes sharp novel views from sparse motion-blurred images. 
DeblurNVS requires no per-scene optimization and achieves better visual quality than existing baselines.}
\label{fig:teaser}
\end{figure}






{Novel view synthesis (NVS) is a fundamental problem in computer vision and graphics, aiming to synthesize photorealistic images from unobserved viewpoints given a set of captured views~\cite{mildenhall2021nerf,kerbl3Dgaussians,yu2024viewcrafter,ding2025futuregs,shi2026sparse4dgs}.} 
{It is a core component of autonomous driving~\cite{tian2025drivingforward,chen2025dggt,miao2025evolsplat}, virtual reality~\cite{li2025radiance}, embodied perception~\cite{huang2025enerverse,chen2025deepverse,zhu2025aether}, and 3D content creation~\cite{ye2024gaussian,wang2024gaussianeditor,shi2025realm}.} 
{Recent progress in neural radiance fields (NeRF)~\cite{mildenhall2021nerf} and 3D Gaussian Splatting (3DGS)~\cite{kerbl3Dgaussians} has substantially improved reconstruction quality and rendering efficiency.} 
{More recently, generative NVS methods have introduced diffusion priors to synthesize novel views implicitly, improving visual fidelity in sparse-view settings~\cite{yu2024viewcrafter,jang2026repurposing}.}
{Despite these advances, most NVS methods are still developed under clean-observation assumptions.} 
{They typically rely on sharp image structures and reliable cross-view geometric cues, both of which can be degraded in practical capture.} 
{Motion blur, caused by camera shake, scene motion, or finite exposure, corrupts local details and weakens multi-view correspondences~\cite{ma2022deblur,wang2023bad,lee2024sharp,zhao2024bad,peng2024bags}.} 
{As a result, NVS methods designed for clean inputs may produce unstable geometry and degraded novel views under motion-blurred observations.}

{A key requirement for practical sparse-view NVS is to avoid time-consuming scene-specific optimization.} 
{Generalizable reconstruction methods, such as MVSplat~\cite{chen2024mvsplat} and AnySplat~\cite{jiang2025anysplat}, improve efficiency by predicting 3D Gaussian primitives directly from sparse inputs.} 
{However, their reconstruction-and-rendering pipeline remains strongly tied to the observed image evidence and the recovered geometry, making it vulnerable when blur has already damaged image structures and correspondence cues.} 
{Generative NVS methods, including MVSplat360~\cite{chen2024mvsplat360}, ViewCrafter~\cite{yu2024viewcrafter}, TrajectoryCrafter~\cite{yu2025trajectorycrafter}, and GLD~\cite{jang2026repurposing}, further leverage diffusion priors to improve visual quality and infer plausible unseen content.} 
{Nevertheless, these models are primarily designed for clean inputs and do not explicitly address the representation ambiguity introduced by motion blur.}

{Blur-aware NVS methods explicitly model the blur formation process and can recover sharp scene representations from degraded observations.} 
{However, they typically rely on costly per-scene optimization, which limits their applicability to efficient and generalizable sparse-view synthesis.} 
{Another possible workaround is to deblur each input image with an off-the-shelf 2D restoration model before applying a standard NVS method.} 
{This cascaded strategy is also problematic because image deblurring is performed independently for each view and does not enforce multi-view geometric consistency.} 
{The restored views may therefore contain view-dependent artifacts, inconsistent textures, or distorted local structures, which can further mislead the subsequent NVS model.}

{In this work, we propose DeblurNVS, a generalizable framework for synthesizing sharp novel views from sparse motion-blurred images without per-scene optimization, as shown in Fig.~\ref{fig:teaser}.} 
{The central idea is to restore the intermediate geometric representations required for multi-view reasoning, so that blurred inputs can provide reliable structure and correspondence cues before target-view synthesis.} 
{DeblurNVS implements this idea through a multi-stage latent learning framework: a context restoration module recovers sharp geometric latents from blurred observations, 
a target latent synthesis stage generates novel-view latents from the restored context and target camera pose via latent diffusion, 
and a lightweight decoder reconstructs the final RGB image.} 
{By addressing blur in a geometry-aware latent space, DeblurNVS avoids the view-inconsistent artifacts of independent 2D deblurring and provides more reliable guidance for sparse-view NVS.}

{Training such a generalizable deblur-aware NVS model requires large-scale scene-level supervision with motion-blurred inputs and sharp target views.} 
{Existing deblurring datasets are mainly designed for 2D image or video restoration, while available deblur-aware NVS datasets contain too few scenes to support end-to-end generalizable training.} 
{As summarized in Tab.~\ref{tab:dataset_comparison_intro}, GoPro~\cite{nah2017deep}, HIDE~\cite{shen2019human}, RealBlur~\cite{rim2020real}, and REDS~\cite{nah2019ntire} provide useful blur supervision but limited scene diversity compared with DL3DV-10K.} 
{Deblur-NeRF~\cite{ma2022deblur}, meanwhile, contains only a small number of NVS scenes and is mainly used for per-scene evaluation.} 
{To support large-scale training, we construct a motion-blurred NVS dataset from DL3DV-10K~\cite{ling2024dl3dv} by synthesizing finite-exposure blur through frame interpolation and temporal averaging.}

\begin{table}[t]
\centering
\caption{\textbf{Comparison with existing datasets.} We categorize each dataset by its task, number of scenes, and number of sharp--blur pairs.}
\label{tab:dataset_comparison_intro}
\small
\begin{tabular}{lccc}
\toprule
Name & Task & \# Scenes & \# Pairs \\
\midrule
GoPro~\cite{nah2017deep} & Image Deblur & 22 & 3,214 \\
HIDE~\cite{shen2019human} & Image Deblur & 31 & 8,422 \\
RealBlur~\cite{rim2020real} & Image Deblur & 232 & 4,738 \\
REDS~\cite{nah2019ntire} & Video Deblur & 300 & $\sim$30K \\
Deblur-NeRF~\cite{ma2022deblur} & Deblur NVS & 15 & -- \\
\midrule
DL3DV-10K-Blur (Ours) & Deblur NVS & $\sim$10K & $\sim$5M \\
\bottomrule
\end{tabular}
\end{table}

\vspace{+4pt}
{Our contributions are summarized as follows:}
\begin{itemize}
    \item \textbf{A generalizable formulation for deblur-aware NVS.} 
    {We introduce DeblurNVS, a framework that synthesizes sharp novel views directly from sparse motion-blurred inputs without requiring per-scene optimization.}

    \item \textbf{A multi-stage geometric latent learning strategy.} 
    {We address motion blur in a geometry-aware latent space by restoring context representations before synthesizing target-view latents with camera-aware diffusion.}

    \item \textbf{A large-scale motion-blurred NVS dataset.} 
    {We construct DL3DV-10K-Blur from DL3DV-10K by simulating finite-exposure blur with frame interpolation and temporal averaging, providing large-scale supervision for end-to-end DeblurNVS training.}

    \item \textbf{Comprehensive evaluation under motion blur.} 
    {We evaluate DeblurNVS on real-world and synthetic benchmarks, demonstrating superior perceptual quality and strong generalization compared with existing generalizable NVS baselines.}
    
\end{itemize}

\begin{figure*}
\centering
\includegraphics[width=\textwidth]{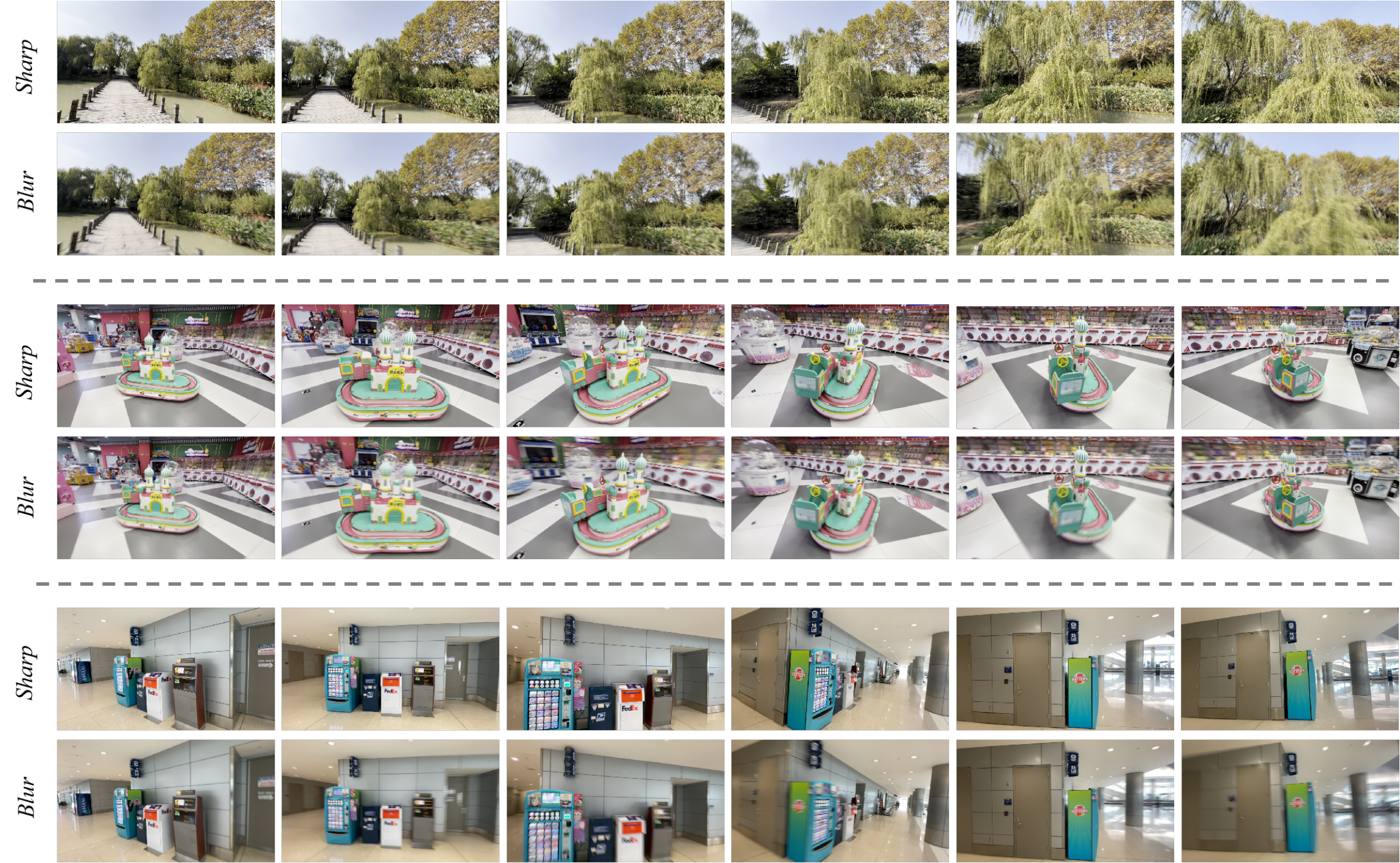}
\caption{\textbf{Visualization of DL3DV-10K-Blur.} We show examples of preprocessed sharp--blur image pairs from our dataset.}
\label{fig:vis_sharp_blur}
\end{figure*}

\section{Related Work}
\subsection{Novel View Synthesis}
Novel view synthesis (NVS) aims to synthesize realistic images from unseen viewpoints based on a set of captured views. Early radiance-field-based methods represent a scene as a continuous volumetric function and synthesize novel views through differentiable volume rendering~\cite{mildenhall2021nerf}. Subsequent works improve rendering quality, training efficiency, and anti-aliasing behavior~\cite{muller2022instant,chen2022tensorf,barron2021mip}.
More recent methods adopt explicit scene representations such as 3D Gaussian splatting (3DGS)~\cite{kerbl3Dgaussians}, significantly improving training and rendering efficiency while maintaining high visual quality. These methods can achieve strong performance, but they require time-consuming per-scene optimization and dense image inputs.

Feed-forward NVS methods avoid scene-specific optimization by learning a direct mapping from input views to novel views or renderable scene representations~\cite{wiles2020synsin,rockwell2021pixelsynth,park2024bridging,zhou2018stereo}. This is often achieved using CNN/Transformer-based architectures~\cite{vaswani2017attention}. Early works such as PixelNeRF~\cite{yu2021pixelnerf} extend radiance fields to the generalizable setting by conditioning neural rendering on image features. Later methods such as PixelSplat~\cite{charatan2024pixelsplat} and MVSplat~\cite{chen2024mvsplat} move to feed-forward prediction of 3D Gaussian primitives, improving both efficiency and generalization.
More recently, vision foundation models~\cite{wang2025vggt,wang2025pi,wang2025depth,shen2025fastvggt} have pushed this direction further. Methods such as AnySplat~\cite{jiang2025anysplat} use stronger geometric priors from large-scale pretraining for feed-forward novel view synthesis. Newer approaches such as Depth Anything 3 (DA3)~\cite{lin2025depth} further strengthen this paradigm by leveraging strong priors from DINOv2~\cite{oquab2023dinov2} backbone.

The rapid progress of diffusion models~\cite{ho2020denoising,song2020denoising,rombach2022high} has shown strong capability in high-quality image synthesis, and has motivated their use in novel view synthesis from single or sparse inputs. Early diffusion-based methods either optimize 3D representations under text-to-image diffusion priors or train pose-conditioned diffusion models for view generation, as in GeNVS~\cite{chan2023generative}, Zero-1-to-3~\cite{liu2023zero}, ZeroNVS~\cite{sargent2024zeronvs}, and Reconfusion~\cite{wu2024reconfusion}. More recent works introduce video diffusion priors to improve view consistency and camera controllability. In particular, ViewCrafter~\cite{yu2024viewcrafter} combines coarse point-based geometry with video diffusion for high-fidelity novel view synthesis, while TrajectoryCrafter~\cite{yu2025trajectorycrafter} further explores diffusion-based camera trajectory control. Recently, Geometric Latent Diffusion (GLD)~\cite{jang2026repurposing} leverages the geometrically consistent feature space of geometric foundation models~\cite{lin2025depth} for multi-view diffusion, further improving cross-view consistency. However, these methods are mainly designed for clean sparse inputs and do not explicitly handle motion-blurred observations.

\subsection{3D Low-Level Vision}
Low-level vision aims to restore degraded visual observations~\cite{mao2016image}, such as blurred~\cite{nah2017deep}, noisy~\cite{zhang2017beyond}, low-resolution~\cite{lim2017enhanced}, or incomplete images. While traditional low-level vision methods are mostly formulated in the 2D image domain, many real-world degradations are inherently coupled with scene geometry, camera motion, and multi-view consistency. This has motivated increasing interest in 3D low-level vision~\cite{liu2026ntire,kwon2025r3evision,xu2025breaking,yu2025evagaussians}, where the goal is not only to recover visually pleasing images but also to reconstruct geometrically consistent 3D scenes or render novel view images.

Existing methods model image degradations during differentiable rendering, enabling reconstruction from noisy, low-light, blurred, weathered, or low-resolution observations~\cite{jiang2026denoisesplat,feng2024srgs,wang2023bad,peng2024bags,zhao2024bad}. However, most existing methods still rely on per-scene optimization. Some recent works attempt to address this limitation with feed-forward reconstruction frameworks~\cite{hu2026srsplat,jiang2026denoisesplat,feng2026sr3r}. Nevertheless, degraded inputs often weaken reliable visual correspondences across views, making it difficult to aggregate multi-view features and render high-fidelity novel view results.

\subsection{Motion Deblurring \& NVS from Motion-Blurred Images}

Motion deblurring seeks to recover latent sharp images from blurred observations caused by camera or object motion during exposure~\cite{jin2018learning}. Existing methods have made substantial progress in single-image~\cite{nah2017deep, tao2018scale,kupyn2019deblurgan} and video-based~\cite{su2017deep, pan2020cascaded, gao2019dynamic} settings. Early learning-based approaches mainly rely on CNNs to restore sharp images from blurred inputs~\cite{tao2018scale}, while later works introduce stronger Transformer-based architectures to improve restoration quality and efficiency~\cite{zamir2022restormer}. More recently, diffusion-based methods further improve perceptual quality by exploiting stronger generative priors~\cite{kong2025deblurdiff}. Despite their strong performance, these methods are primarily developed for 2D image restoration and focus on recovering sharp observations in the input views. 

Recent works have begun to address novel view synthesis directly from motion-blurred inputs. Deblur-NeRF~\cite{ma2022deblur} is an early method that recovers a sharp radiance field from blurry images by explicitly simulating the blur formation process during optimization. BAD-NeRF~\cite{wang2023bad} further extends this direction by jointly modeling motion blur and inaccurate camera poses, and recovering camera motion trajectories during exposure time. With the rise of explicit 3D scene representations, BAGS~\cite{peng2024bags} introduces additional 2D blur modeling capacities into Gaussian Splatting to reconstruct 3D-consistent scenes under image-wise blur, while BAD-Gaussians~\cite{zhao2024bad} adopts a bundle-adjusted Gaussian representation to jointly handle severe motion blur and pose errors. Although these methods achieve strong NVS performance under blurred inputs, they require computationally expensive per-scene optimization.

\begin{figure*}
\centering
\includegraphics[width=\textwidth]{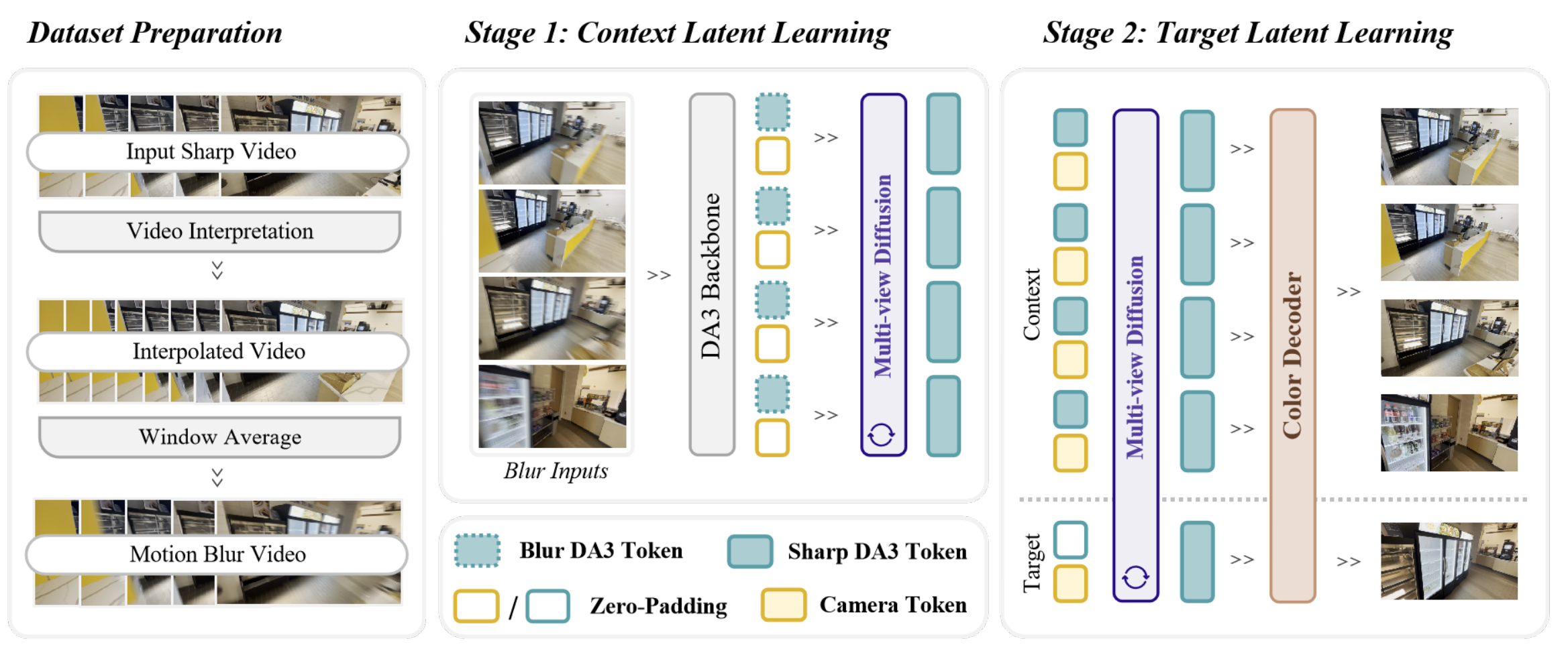}
\caption{\textbf{Pipeline of DeblurNVS.} 
1) We construct a large-scale motion-blurred dataset from DL3DV-10K via frame interpolation and temporal averaging. 
2) Given motion-blurred inputs, DeblurNVS learns sharp multi-view latents by jointly adapting the DA3 backbone and multi-view diffusion. 
3) Camera tokens are further introduced to guide novel-view latent synthesis, and a shared color decoder maps the restored and synthesized latents to sharp RGB images.}
\label{fig1}
\end{figure*}

\section{Method of DeblurNVS}
\subsection{Preliminary: Geometric Latent Diffusion (GLD)}
Recent diffusion-based NVS methods often leverage external geometry conditioning, such as depth-based warping~\cite{cao2025mvgenmaster,kwak2025aligned,seo2024genwarp}. These approaches rely on latent spaces originally designed for single-image synthesis models, which often result in multi-view inconsistency. 
To tackle this challenge, GLD~\cite{jang2026repurposing} proposes to repurpose the feature space of geometric foundation models (e.g., DA3~\cite{lin2025depth}) as the latent space for multi-view diffusion. Given a set of context views $\mathcal{I}_s = \{\mathbf{I}_i\}_{i=1}^K$, GLD uses a pretrained DA3 encoder $E_{\mathrm{DA3}}$ to extract {latents}:
\begin{equation}
\mathcal{Z}_c = E_{\mathrm{DA3}}(\mathcal{I}_c),
\qquad
\mathcal{I}_c \in \mathbb{R}^{K\times 3 \times H \times W},
\end{equation}
where $\mathcal{Z}_c=\{\mathbf{z}_i\}_{i=1}^K$ denotes the DA3 {latents} of the context views,
while the target view is represented by its DA3 latent
\begin{equation}
\mathbf{z}^{\star} = E_{\mathrm{DA3}}(\mathbf{I}^{\star}).
\end{equation}

During training, GLD perturbs $\mathbf{z}^{\star}$ with Gaussian noise:
\begin{equation}
\mathbf{z}^{\star}_t = \alpha_t \mathbf{z}^{\star} + \sigma_t \boldsymbol{\epsilon},
\qquad
\boldsymbol{\epsilon}\sim\mathcal{N}(0,\mathbf{I}),
\end{equation}
and learns a denoising model that predicts the noise conditioned on the context latent set $\mathcal{Z}_s$ and camera latent $\mathbf{c}$:
\begin{equation}
\mathcal{L}_{\mathrm{GLD}}
=
\mathbb{E}_{\mathbf{z}^{\star},\boldsymbol{\epsilon},t}
\left[
\left\|
\boldsymbol{\epsilon}
-
\epsilon_\theta(\mathbf{z}^{\star}_t,t,\mathcal{Z}_,\mathbf{c})
\right\|_2^2
\right].
\end{equation}

In this way, GLD performs novel view synthesis by generating the target-view representation in the DA3 latent space, and the predicted latent is decoded into the final RGB novel view using a lightweight ViT decoder.
{DeblurNVS adopts this latent-space formulation and adapts it to sparse motion-blurred inputs.}

\subsection{Motion-Blurred Dataset Preparation}
In real-world imaging, motion blur arises from the finite exposure time of a camera. Instead of capturing an instantaneous sharp image, the sensor integrates the incoming radiance over the exposure interval. Therefore, a blurry image can be modeled as the temporal integration of latent sharp images~\cite{jin2018learning}:
\begin{equation}
\mathbf{I}^{\mathrm{blur}} = \frac{1}{T}\int_{t_0}^{t_0+T} \mathbf{I}^{\mathrm{sharp}}(t)\, dt,
\end{equation}
where \(I^{\mathrm{sharp}}(t)\) denotes the latent sharp image at time \(t\), and \(T\) is the exposure duration.

To simulate this blur formation process, we first interpolate DL3DV-10K~\cite{ling2024dl3dv} to obtain temporally denser frames using a video interpolation model~\cite{niklaus2021revisiting}. We then approximate the exposure integral by averaging a temporal window of interpolated frames. This process can be formulated as:
\begin{equation}
\mathbf{I}^{\mathrm{blur}} \approx \frac{1}{N}\sum_{i=1}^{N} \mathbf{I}_i^{\mathrm{sharp}},\quad N \sim \mathcal{U}\{5,7,9,11\},
\end{equation}
where \(\{I_i^{\mathrm{sharp}}\}_{i=1}^{N}\) are the interpolated sharp frames within the sampled window. We randomly sample the window size \(N\) from \(\{5,7,9,11\}\) to {increase the diversity of blur patterns. We present examples of our preprocessed dataset in Fig.~\ref{fig:vis_sharp_blur}.}

\subsection{Multi-Stage Latent Learning}

Vanilla GLD is designed for sharp input images, where DA3 can extract high-quality latent representations for subsequent novel view synthesis. However, under motion blur, the pretrained DA3 encoder fails to produce reliable latent representations, leading to degraded NVS quality.
To address this issue, we decompose this task into two stages: \emph{context latent learning} and \emph{target latent learning}. 

\vspace{+6pt}
\noindent\textit{\textbf{Context Latent Learning.}} In this stage, we aim to recover sharp DA3 latents from motion-blurred images without introducing any camera conditioning. Given a set of motion-blurred images \(\mathcal{I}_c^{\mathrm{blur}}=\{\mathbf{I}_i^{\mathrm{blur}}\}_{i=1}^{K}\), we first feed them into a student DA3 encoder equipped with lightweight LoRA adapters~\cite{hu2022lora}:
\begin{equation}
\widetilde{\mathcal{Z}}_c
=
E_{\mathrm{DA3}}^{\mathrm{LoRA}}(\mathcal{I}_c^{\mathrm{blur}}),
\qquad
\widetilde{\mathcal{Z}}_c = \{\tilde{\mathbf{z}}_i\}_{i=1}^{K}.
\end{equation}

Based on the blurred context {latents} \(\widetilde{\mathcal{Z}}_c\), we train a context latent diffusion model to recover sharp context representations. Following GLD, we formulate this stage as a latent denoising process. Specifically, the target sharp latent is perturbed as
\begin{equation}
\mathcal{Z}_{c,t}^{\mathrm{sharp}}
=
\alpha_t \mathcal{Z}_c^{\mathrm{sharp}}
+
\sigma_t \boldsymbol{\epsilon},
\qquad
\boldsymbol{\epsilon}\sim\mathcal{N}(\mathbf{0},\mathbf{I}),
\end{equation}
and the context diffusion model is optimized with
\begin{equation}
\mathcal{L}_{\mathrm{ctx}}
=
\mathbb{E}_{\mathcal{Z}_c^{\mathrm{sharp}},\boldsymbol{\epsilon},t}
\left[
\left\|
\boldsymbol{\epsilon}
-
\epsilon_{\theta}^{\mathrm{ctx}}
\big(
\mathcal{Z}_{c,t}^{\mathrm{sharp}},
t,
\widetilde{\mathcal{Z}}_c,
\mathbf{0}
\big)
\right\|_2^2
\right].
\end{equation}

Here, the camera condition is explicitly removed by zero-padding the camera embedding, i.e., \(\mathbf{c}=\mathbf{0}\), so that the model focuses purely on appearance restoration rather than geometry-aware view synthesis. The target sharp latent \(\mathcal{Z}_c^{\mathrm{sharp}}\) is extracted by a frozen teacher DA3 encoder from the corresponding sharp context images:
\begin{equation}
\mathcal{Z}_c^{\mathrm{sharp}}
=
E_{\mathrm{DA3}}(\mathcal{I}_c^{\mathrm{sharp}}),
\qquad
\mathcal{Z}_c^{\mathrm{sharp}}=\{\mathbf{z}_i^{\mathrm{sharp}}\}_{i=1}^{K}.
\end{equation}
During this stage, the LoRA parameters in the student DA3 encoder and the context diffusion model are jointly optimized, while the teacher DA3 encoder remains frozen.

\begin{table}[t]
\centering
\caption{Comparison on DL3DV-Bench with 3 input views. Best results are highlighted in \colorbox{bestred}{light red}.}
\label{tab:dl3dv_bench}
\small
\begin{adjustbox}{max width=0.95\linewidth}
\begin{tabular}{lcccccc}
\toprule
Method & PSNR$\uparrow$ & SSIM$\uparrow$ & LPIPS$\downarrow$ & DISTS$\downarrow$ & FID$\downarrow$ & Time$\downarrow$  \\
\midrule
3DGS & 14.769& 0.527& 0.422& 0.262& 233.40& 32 min\\
BAGS & 15.184&\cellcolor{bestred}0.539&0.414& 0.259& 221.76&12 min\\
DA3 & 10.945& 0.357& 0.581& 0.337& 275.52&\cellcolor{bestred}0.03 s\\
GLD & 11.461& 0.385& 0.503& 0.235& 150.90&9.76 s\\
\midrule
Ours & \cellcolor{bestred}15.549& 0.441& \cellcolor{bestred}0.367&\cellcolor{bestred} 0.174& \cellcolor{bestred}101.36&0.60 s\\

\bottomrule
\end{tabular}
\end{adjustbox}
\end{table}

\begin{table*}[t]
\centering
\caption{Quantitative comparison on DeblurNeRF-Real dataset. We additionally report runtime, where scene-specific methods are measured by average per-scene training time in minutes, and {generalizable methods are measured by average per-novel-view inference time in seconds}. Best results are highlighted in \colorbox{bestred}{light red}.}
\label{tab:real_main_grouped}
\small
\begin{adjustbox}{max width=0.98\textwidth}
\begin{tabular*}{\textwidth}{@{\extracolsep{\fill}} c c c c c c c c c}
\toprule
\textbf{Views} & \textbf{Type} & \textbf{Method} & \textbf{PSNR$\uparrow$} & \textbf{SSIM$\uparrow$} & \textbf{LPIPS$\downarrow$} & \textbf{DISTS$\downarrow$} & \textbf{FID$\downarrow$} & \textbf{Time$\downarrow$} \\
\midrule
\multirow{5}{*}{3}
& \multirow{2}{*}{Scene-specific}
& 3DGS & 18.454 & 0.512 & 0.400 & 0.253 & 166.348 & 12 min \\
& & BAGS & \cellcolor{bestred}18.813 & \cellcolor{bestred}0.527 & 0.385 & 0.241 & 151.080 & 30 min \\
\cmidrule(lr){2-9}
& \multirow{3}{*}{Generalizable}
& DA3 & 13.610 & 0.356 & 0.608 & 0.411 & 306.244 & \cellcolor{bestred}0.03 s \\
& & GLD & 12.868 & 0.329 & 0.570 & 0.281 & 166.513 & 9.77 s \\
& & \textbf{Ours} & 17.132 & 0.433 & \cellcolor{bestred}0.335 & \cellcolor{bestred}0.142 & \cellcolor{bestred}79.648 & 0.60 s \\
\midrule
\multirow{5}{*}{6}
& \multirow{2}{*}{Scene-specific}
& 3DGS & \cellcolor{bestred}19.554 & \cellcolor{bestred}0.554 & 0.374 & 0.247 & 177.526 & 12 min \\
& & BAGS & 19.541 & 0.544 & 0.330 & 0.201 & 111.186 & 30 min \\
\cmidrule(lr){2-9}
& \multirow{3}{*}{Generalizable}
& DA3 & 13.687 & 0.369 & 0.620 & 0.464 & 349.469 & \cellcolor{bestred}0.05 s \\
& & GLD & 13.638 & 0.354 & 0.538 & 0.276 & 168.222 & 13.91 s \\
& & \textbf{Ours} & 17.893 & 0.464 & \cellcolor{bestred}0.301 & \cellcolor{bestred}0.129 & \cellcolor{bestred}73.571 & 0.80 s \\
\midrule
\multirow{5}{*}{9}
& \multirow{2}{*}{Scene-specific}
& 3DGS & \cellcolor{bestred}19.720 & \cellcolor{bestred}0.560 & 0.370 & 0.243 & 164.571 & 13 min \\
& & BAGS & 18.237 & 0.493 & 0.319 & 0.173 & 91.386 & 32 min \\
\cmidrule(lr){2-9}
& \multirow{3}{*}{Generalizable}
& DA3 & 13.835 & 0.373 & 0.620 & 0.490 & 354.231 & \cellcolor{bestred}0.08 s \\
& & GLD & 13.771 & 0.355 & 0.532 & 0.266 & 151.398 & 18.49 s \\
& & \textbf{Ours} & 18.091 & 0.475 & \cellcolor{bestred}0.290 & \cellcolor{bestred}0.125 & \cellcolor{bestred}70.221 & 1.04 s \\
\bottomrule
\end{tabular*}
\end{adjustbox}
\end{table*}

\begin{figure*}
\centering
\includegraphics[width=\textwidth]{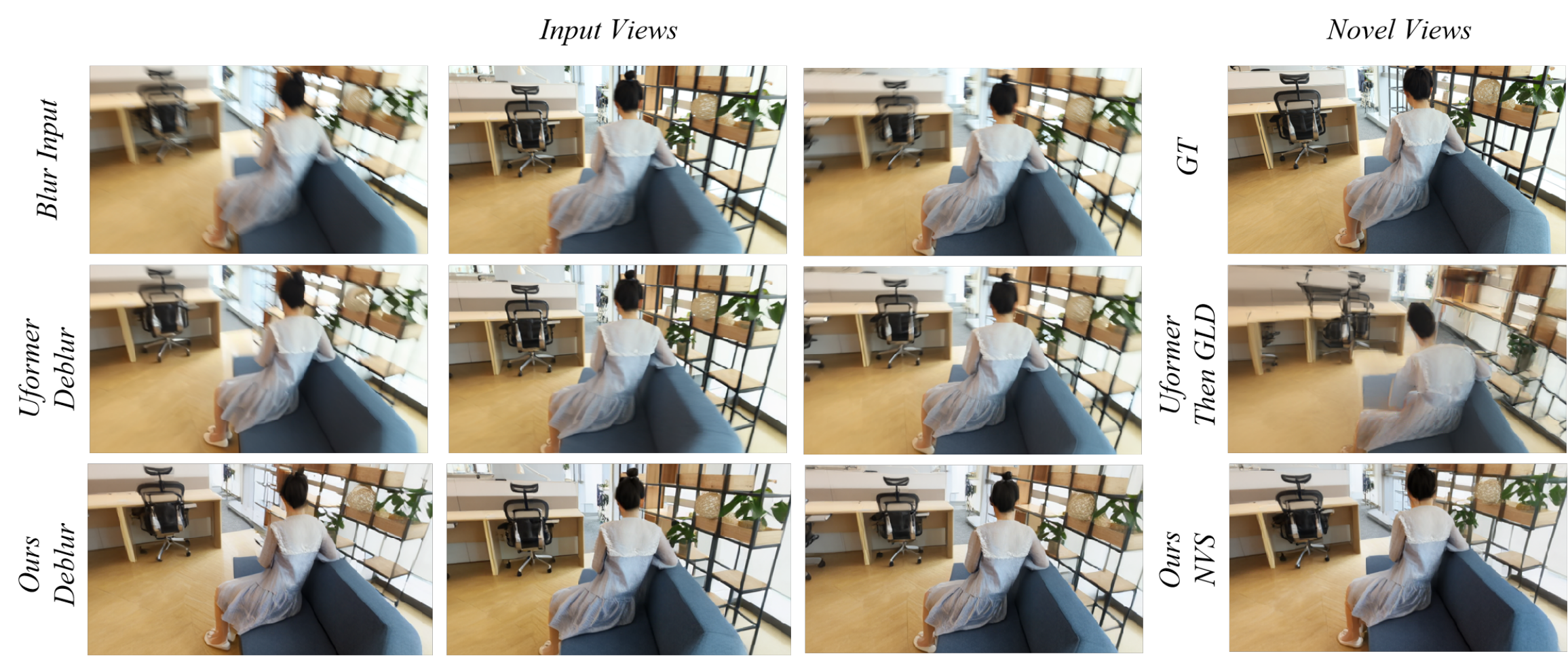}
\caption{\textbf{2D Deblur+GLD vs Ours.} Left: blurred inputs, Uformer-restored inputs, and our restored inputs. 
Right: ground-truth novel view, Uformer+GLD result, and ours. 
Independent 2D deblurring remains limited and multi-view inconsistent, leading to degraded GLD synthesis, while our method produces more consistent restorations and sharper novel views.
}
\label{fig:ablation_iformer}
\end{figure*}

\begin{table*}[t]
\centering
\caption{Quantitative comparison on DeblurNeRF-Blender dataset. We additionally report runtime, where scene-specific methods are measured by average per-scene training time in minutes, and {generalizable methods are measured by average per-novel-view inference time in seconds}. Best results are highlighted in \colorbox{bestred}{light red}.}
\label{tab:blender_main_grouped}
\small
\begin{adjustbox}{max width=0.98\textwidth}
\begin{tabular*}{\textwidth}{@{\extracolsep{\fill}} c c c c c c c c c}
\toprule
\textbf{Views} & \textbf{Type} & \textbf{Method} & \textbf{PSNR$\uparrow$} & \textbf{SSIM$\uparrow$} & \textbf{LPIPS$\downarrow$} & \textbf{DISTS$\downarrow$} & \textbf{FID$\downarrow$} & \textbf{Time$\downarrow$} \\
\midrule
\multirow{5}{*}{3}
& \multirow{2}{*}{Scene-specific}
& 3DGS & 17.873 & 0.482 & 0.406 & 0.249 & 177.976 & 12 min \\
& & BAGS & \cellcolor{bestred}18.021 & \cellcolor{bestred}0.485 & 0.404 & 0.247 & 173.374 & 30 min \\
\cmidrule(lr){2-9}
& \multirow{3}{*}{Generalizable}
& DA3 & 11.745 & 0.282 & 0.665 & 0.456 & 344.404 & \cellcolor{bestred}0.04 s \\
& & GLD & 16.009 & 0.403 & 0.451 & 0.243 & 162.288 & 10.07 s \\
& & \textbf{Ours} & 16.049 & 0.387 & \cellcolor{bestred}0.340 & \cellcolor{bestred}0.149 & \cellcolor{bestred}103.528 & 0.66 s \\
\midrule
\multirow{5}{*}{6}
& \multirow{2}{*}{Scene-specific}
& 3DGS & 19.177 & 0.531 & 0.378 & 0.238 & 155.986 & 12 min \\
& & BAGS & \cellcolor{bestred}19.903 & \cellcolor{bestred}0.565 & 0.330 & 0.203 & 126.352 & 30 min \\
\cmidrule(lr){2-9}
& \multirow{3}{*}{Generalizable}
& DA3 & 11.315 & 0.281 & 0.681 & 0.479 & 409.778 & \cellcolor{bestred}0.06 s \\
& & GLD & 17.287 & 0.454 & 0.401 & 0.229 & 139.073 & 14.72 s \\
& & \textbf{Ours} & 16.385 & 0.412 & \cellcolor{bestred}0.317 & \cellcolor{bestred}0.137 & \cellcolor{bestred}81.380 & 0.85 s \\
\midrule
\multirow{5}{*}{9}
& \multirow{2}{*}{Scene-specific}
& 3DGS & 19.857 & 0.540 & 0.366 & 0.233 & 159.834 & 12 min \\
& & BAGS & \cellcolor{bestred}20.629 & \cellcolor{bestred}0.568 & 0.310 & 0.188 & 127.382 & 30 min \\
\cmidrule(lr){2-9}
& \multirow{3}{*}{Generalizable}
& DA3 & 12.250 & 0.304 & 0.656 & 0.492 & 379.662 & \cellcolor{bestred}0.09 s \\
& & GLD & 17.279 & 0.443 & 0.411 & 0.230 & 143.235 & 20.03 s \\
& & \textbf{Ours} & 16.481 & 0.426 & \cellcolor{bestred}0.306 & \cellcolor{bestred}0.132 & \cellcolor{bestred}82.336 & 1.13 s \\
\bottomrule
\end{tabular*}
\end{adjustbox}
\end{table*}

\vspace{+6pt}
\noindent \textbf{\textit{Target Latent Learning.}}
After obtaining the restored sharp context {latents} \(\hat{\mathcal{Z}}_c\) from the first stage, we further learn to synthesize the latent representation of the target novel views. Unlike the context stage, this stage explicitly introduces {camera pose conditioning}. During training, given a sharp target view \(\mathbf{I}^{\star,\mathrm{sharp}}\), we use the frozen teacher DA3 encoder to extract its target latent:
\begin{equation}
\mathbf{z}^{\star,\mathrm{sharp}}
=
E_{\mathrm{DA3}}(\mathbf{I}^{\star,\mathrm{sharp}}).
\end{equation}

Meanwhile, we use the frozen teacher DA3 model to predict the camera parameters \(\mathbf{c}\) for both context and target views. {In our implementation}, all camera parameters are kept in the native DA3 coordinate system.
Following GLD, we perturb the target sharp latent with Gaussian noise:
\begin{equation}
\mathbf{z}^{\star,\mathrm{sharp}}_t
=
\alpha_t \mathbf{z}^{\star,\mathrm{sharp}}
+
\sigma_t \boldsymbol{\epsilon},
\qquad
\boldsymbol{\epsilon}\sim\mathcal{N}(\mathbf{0},\mathbf{I}),
\end{equation}
and train a target latent diffusion model conditioned on the restored context latent \(\hat{\mathcal{Z}}_c\) and the camera condition \(\mathbf{c}\):
\begin{equation}
\mathcal{L}_{\mathrm{tgt}}
=
\mathbb{E}_{\mathbf{z}^{\star,\mathrm{sharp}},\boldsymbol{\epsilon},t}
\left[
\left\|
\boldsymbol{\epsilon}
-
\epsilon_{\theta}^{\mathrm{tgt}}
\big(
\mathbf{z}^{\star,\mathrm{sharp}}_t,
t,
\hat{\mathcal{Z}}_c,
\mathbf{c}
\big)
\right\|_2^2
\right].
\end{equation}

Unlike the first stage, which focuses on camera-free restoration of observed context latents, the second stage targets geometry-aware synthesis of unseen target latents. Accordingly, the diffusion model is now conditioned on both the restored context latent \(\hat{\mathcal{Z}}_c\) and the camera parameters \(\mathbf{c}\).

To obtain the final RGB outputs, we train a lightweight decoder \(D_{\mathrm{rgb}}\) in this stage. It takes the restored context latents and the synthesized target latent as input, and is supervised by the corresponding sharp images with pixel, LPIPS perceptual~\cite{zhang2018unreasonable}, and adversarial~\cite{goodfellow2014generative} losses:
\begin{equation}
\mathcal{L}_{\mathrm{rgb}}
=
\lambda_{1}\mathcal{L}_{1}
+
\lambda_{p}\mathcal{L}_{\mathrm{LPIPS}}
+
\lambda_{g}\mathcal{L}_{\mathrm{GAN}}.
\end{equation}

In this way, DeblurNVS enables high-quality novel view synthesis directly from motion-blurred images.

\begin{figure*}
\centering
\includegraphics[width=\textwidth]{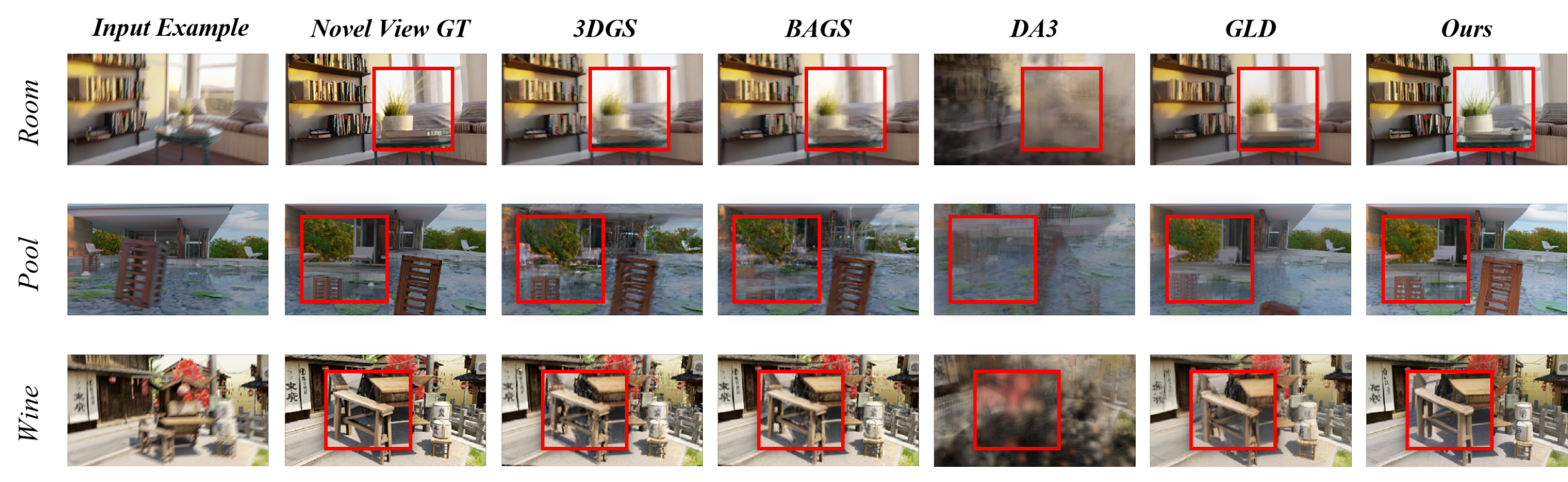}
\caption{\textbf{Qualitative results on the DeblurNeRF-Blender dataset.} 
Compared with baseline methods, our method reconstructs sharper details and achieves higher visual fidelity under motion-blurred inputs.}
\label{fig:main_comparison_blender}
\end{figure*}

\begin{figure*}
\centering
\includegraphics[width=\textwidth]{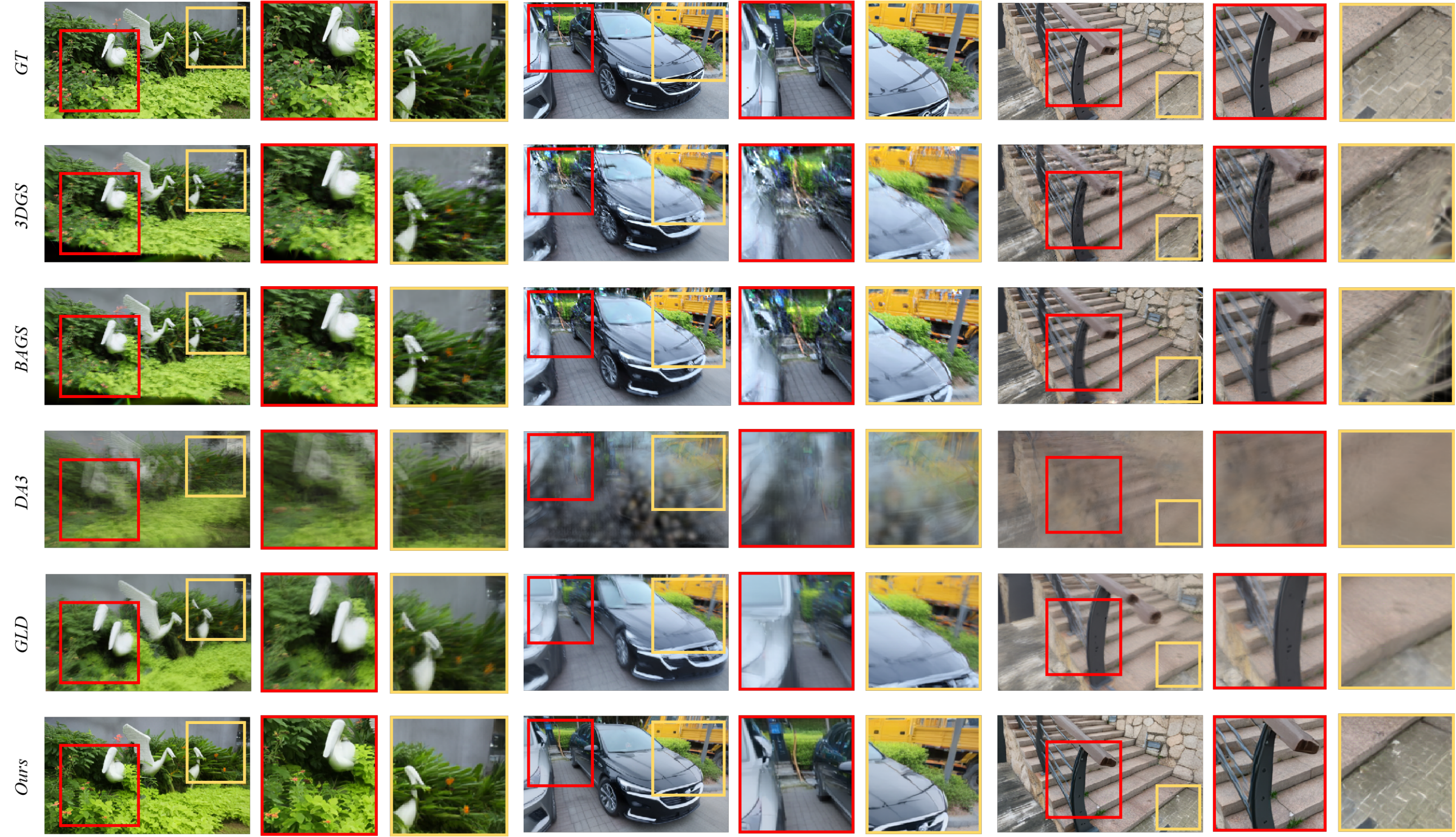}
\caption{\textbf{Qualitative results on the DeblurNeRF-Real dataset.} 
Compared with baseline methods, our method reconstructs sharper details and achieves higher visual fidelity under motion-blurred inputs.}
\label{fig: main_comparison}
\end{figure*}

\section{Experiments}

\subsection{Settings}



\noindent \textit{\textbf{Datasets}} 
For training, we build a large-scale {motion-blurred} dataset based on the full DL3DV-10K~\cite{ling2024dl3dv}, {which comprises} around 10,000 scenes, each with roughly 500 images.
For evaluation, we use three benchmarks:
1) Following DeblurNeRF~\cite{ma2022deblur}, we evaluate on 10 real-world captured scenes, where the context views are motion-blurred and the target views are sharp. 2) We further evaluate on five synthetic Blender scenes, where motion blur is simulated by perturbing camera poses and averaging renders from interpolated poses. 3) Moreover, we select a subset from DL3DV-Bench and synthesize motion-blurred input views for evaluation.

\vspace{+4pt}
\noindent \textit{\textbf{Baselines.}} 
\textit{1) Per-scene optimized methods:} For this category, we compare against Vanilla 3DGS~\cite{kerbl3Dgaussians} and BAGS~\cite{peng2024bags}, the latter of which is specifically designed for blurred inputs. 
\textit{2) Generative methods:} We further compare against DA3~\cite{lin2025depth}, a state-of-the-art feed-forward NVS approach, and GLD~\cite{jang2026repurposing}, a state-of-the-art diffusion-based NVS model.

\vspace{+4pt}
\noindent \textit{\textbf{Metrics.}} 
We report PSNR and SSIM~\cite{wang2004image} to measure pixel-wise similarity. 
To evaluate perceptual quality, we further report LPIPS~\cite{zhang2018unreasonable}, DISTS~\cite{ding2022image}, and FID~\cite{heusel2017gans}, which better reflect visual realism and feature-level consistency.

\vspace{+4pt}
\noindent \textit{\textbf{Implementation Details. }}We build DeblurNVS on the GLD backbone. We replace the original level-wise cascade with a single-level DA3 latent representation for more efficient training and inference. At inference, the NVS diffusion model is sampled with Euler integration for 12 denoising steps, while the context refiner diffusion model uses 8 steps. Experiments are conducted at a resolution of {\(280 \times 504\)}. The overall training time is about {2--3} days.  For decoder training, the RGB loss uses $\lambda_{1}=1.0$, $\lambda_{p}=1.0$, and $\lambda_{g}=0.75$.  Results can be reproduced on a single 48 GB NVIDIA 4090 GPU.

\subsection{Main Results}

\noindent \textit{\textbf{Quantitative Comparisons.}} 
We present quantitative results in Tab.~\ref{tab:dl3dv_bench}, Tab.~\ref{tab:real_main_grouped}, and Tab.~\ref{tab:blender_main_grouped}. 
Scene-specific optimization methods generally obtain higher PSNR and SSIM, since they are optimized separately for each test scene. 
In contrast, DeblurNVS {performs inference on unseen scenes without per-scene optimization}. 
Moreover, PSNR and SSIM are distortion-oriented metrics and may favor smoother predictions in deblurring scenarios, making them less aligned with perceptual sharpness. 
Despite being slightly lower on these metrics, DeblurNVS consistently improves perceptual metrics such as LPIPS, DISTS, and FID, indicating better visual realism and sharper reconstruction.

\noindent \textit{\textbf{Qualitative Comparisons.}} 
As shown in Fig.~\ref{fig:main_comparison_blender} and Fig.~\ref{fig: main_comparison}, scene-specific methods such as 3DGS and BAGS often produce floating artifacts under motion-blurred inputs. 
Meanwhile, generalizable baselines such as DA3 and GLD, especially DA3, suffer from severe geometric misalignment due to unreliable correspondences extracted from blurred observations. 
In contrast, DeblurNVS reconstructs more coherent geometry and sharper visual details, leading to more realistic and structurally consistent novel views.

\noindent\textbf{\textit{Inference Time.}} In terms of efficiency, scene-specific methods require tens of minutes of optimization for each test scene, while generalizable methods render novel views {without scene-specific optimization at test time}. 
DeblurNVS is much faster than GLD because we simplify the diffusion framework by reducing the number of denoising steps and removing cascaded operations across DA3 feature levels. 
However, as a diffusion-based method, DeblurNVS is naturally slower than fully feed-forward models such as DA3.

\begin{figure*}
\centering
\includegraphics[width=\textwidth]{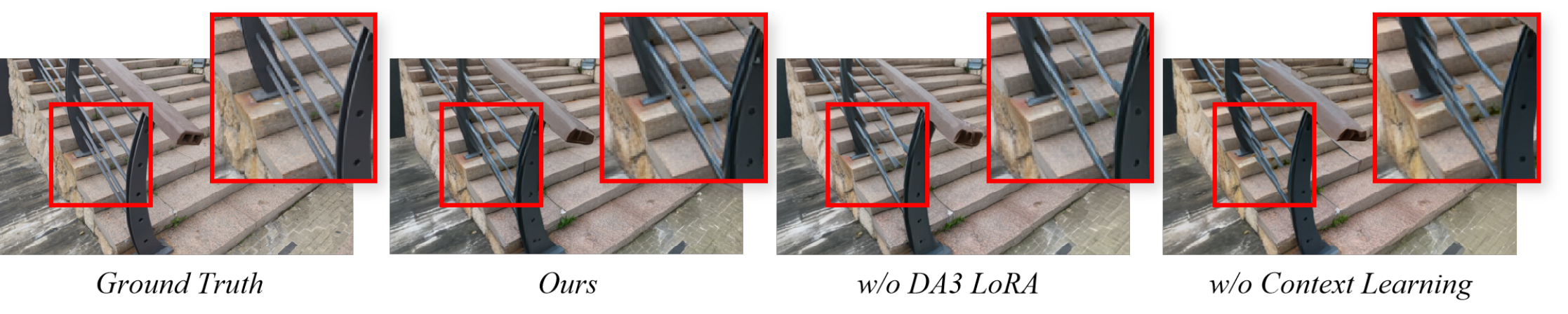}
\caption{\textbf{Ablation study on overall architecture.} 
Removing LoRA adaptation from the DA3 backbone or discarding the first-stage context learning leads to degraded geometric structures.}
\label{fig: arch_ablation}
\end{figure*}

\subsection{Ablation Study}

\noindent \textbf{\textit{Overall Architecture.}} 
We ablate two key design choices in our framework on DeblurNeRF-Real dataset with 3 input views: removing the first-stage context learning, and disabling LoRA adaptation on the DA3 backbone. 
As shown in Tab.~\ref{tab:ablation_overall}, both variants lead to inferior overall performance compared with the full model, validating the effectiveness of each component. We further provide qualitative comparisons of these ablations in Fig.~\ref{fig: arch_ablation}. 
Our full model reconstructs more accurate geometry, whereas removing either design choice makes it more difficult for the model to capture reliable geometric correspondences, leading to structural misalignment.


\begin{figure*}
\centering
\includegraphics[width=\textwidth]{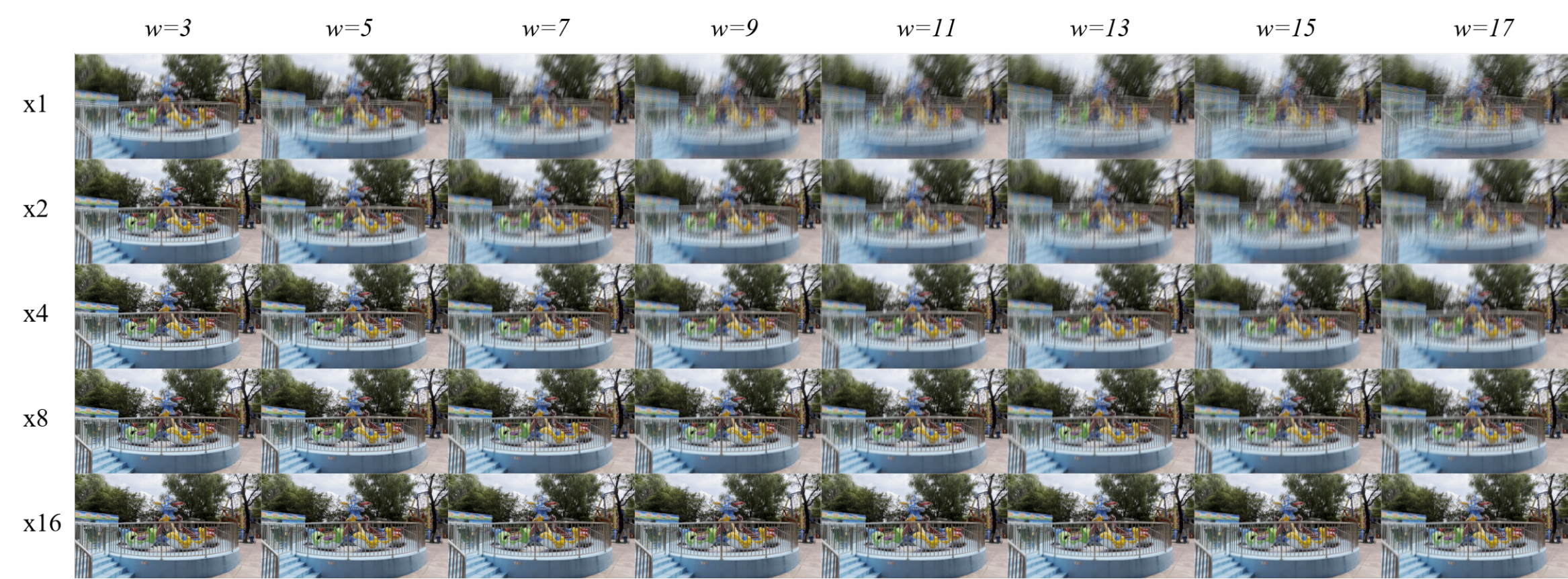}
\caption{\textbf{Ablation on hyperparameters of DL3DV-10K-Blur.} 
Columns correspond to different averaging window sizes, while rows correspond to different frame interpolation rates.}
\label{fig:ablation_motionblur_dataset}
\end{figure*}


\begin{table}[t]
\centering
\caption{\textbf{Ablation study on overall architecture.} CL denotes the first-stage context latent learning, and LoRA denotes the LoRA adaptation applied to the DA3 backbone.}
\label{tab:ablation_overall}
\small
\begin{adjustbox}{max width=0.95\linewidth}
\begin{tabular}{lccccc}
\toprule
\textbf{Method} & \textbf{PSNR$\uparrow$} & \textbf{SSIM$\uparrow$} & \textbf{LPIPS$\downarrow$} & \textbf{DISTS$\downarrow$} & \textbf{FID$\downarrow$} \\
\midrule
w/o CL & 16.989 & \textbf{0.437} & 0.345 & 0.149 & 82.828 \\
w/o LoRA & 17.122 & 0.433 & 0.336 & 0.145 & 83.148 \\
\midrule
Ours & \textbf{17.132} & 0.433 & \textbf{0.335} & \textbf{0.142} & \textbf{79.648} \\
\bottomrule
\end{tabular}
\end{adjustbox}
\end{table}

\noindent \textbf{\textit{2D Deblur+GLD.}} 
We further investigate a cascaded baseline that first restores the blurred input views with an off-the-shelf 2D deblurring model (Uformer~\cite{Wang_2022_CVPR}) and then feeds the restored images into GLD for novel view synthesis. 
The quantitative and qualitative results are shown in Tab.~\ref{tab:ablation_2d_deblur} and Fig.~\ref{fig:ablation_iformer}, respectively. 
As shown in the results, Uformer provides only limited restoration for motion-blurred input views. 
Moreover, since it processes each view independently, the deblurred images still suffer from multi-view inconsistency. 
These residual blur and inconsistent structures make it difficult for GLD to establish reliable cross-view correspondences, leading to degraded NVS performance. 
In contrast, our method outperforms this cascaded baseline in both input-view restoration and novel-view synthesis, demonstrating the advantage of performing blur-aware NVS in a geometry-aware latent space.

\begin{table}[t]
\centering
\caption{{Comparison with the RGB-space cascaded baseline.} Uformer+GLD first deblurs the input views with Uformer and then performs NVS using GLD.}
\label{tab:ablation_2d_deblur}
\small
\begin{adjustbox}{max width=0.95\linewidth}
\begin{tabular}{lccccc}
\toprule
Method & PSNR$\uparrow$ & SSIM$\uparrow$ & LPIPS$\downarrow$ & DISTS$\downarrow$ & FID$\downarrow$ \\
\midrule
GLD & 12.868 & 0.329 & 0.570 & 0.281 & 166.513 \\
Uformer+GLD & 12.643 &0.335&0.602& 0.252 & 151.051\\
\midrule
Ours & \textbf{17.132} & \textbf{0.433 }& \textbf{0.335} & \textbf{0.142} & \textbf{79.648} \\
\bottomrule
\end{tabular}
\end{adjustbox}
\end{table}

\noindent \textbf{\textit{Hyperparameters in DL3DV-10K-Blur.}} 
We visualize motion-blurred images generated with different frame interpolation rates and temporal averaging window sizes in Fig.~\ref{fig:ablation_motionblur_dataset}. 
A low frame interpolation rate provides insufficient temporal samples and results in inaccurate blur simulation. 
In contrast, an excessively high interpolation rate produces overly smooth temporal changes, making the synthesized motion blur less noticeable. 
Therefore, we adopt a moderate interpolation rate of $8\times$ in our dataset construction. 
Under this setting, we randomly sample different temporal averaging window sizes to simulate diverse degrees of motion blur.

\section{Conclusions and Limitations}

We present DeblurNVS, a generalizable framework for sharp novel view synthesis from sparse motion-blurred images. DeblurNVS adopts a multi-stage strategy to decouple multi-view deblurring from novel view synthesis. We further construct DL3DV-10K-Blur to support large-scale end-to-end training. Experiments on synthetic and real-world benchmarks demonstrate that DeblurNVS achieves better perceptual quality and stronger generalization than existing baselines.

Despite its strong visual quality, DeblurNVS still has several limitations. First, as a generative diffusion-based method, it requires iterative denoising during inference and is therefore slower than purely feed-forward NVS models. Second, the generative prior may hallucinate plausible details under severe blur, sparse inputs, or large viewpoint changes. As a result, DeblurNVS can produce visually appealing results that are not strictly consistent with the ground truth, especially in fine textures and occluded regions. Future work may explore faster generative formulations and stronger geometric constraints to improve both efficiency and fidelity.

\clearpage
\newpage

\bibliographystyle{IEEEtran}
\bibliography{main}

\newpage

 




\vfill

\end{document}